\documentclass[11pt]{article}

\usepackage[final]{acl}

\usepackage{times}
\usepackage{latexsym}
\usepackage{multirow}%
\usepackage{amsmath,amssymb,amsfonts}%
\usepackage{amsthm}%
\usepackage[most]{tcolorbox}
\usepackage{mathrsfs}%
\usepackage[title]{appendix}%
\usepackage{xcolor}%
\usepackage{textcomp}%
\usepackage{manyfoot}%
\usepackage{booktabs}%
\usepackage{algorithm}%
\usepackage{color,array}
\usepackage{booktabs}
\usepackage{amsmath}
\usepackage[most]{tcolorbox}
\usepackage{colortbl}  % for \cellcolor
\newcommand{\cmark}{\ding{51}}%
\newcommand{\xmark}{\ding{55}}%
\usepackage{pifont}% http://ctan.org/pkg/pifont
\usepackage{makecell}
\usepackage{xcolor}
\usepackage{arydshln}
\definecolor{chartred}{RGB}{228,36,36}
\definecolor{chartblue}{RGB}{65,105,225}
\definecolor{chartcyan}{RGB}{102,204,204}
\definecolor{chartyellow}{RGB}{255,204,51}
\definecolor{chartgreen}{RGB}{51,153,51}
\usepackage{algorithmicx}%
\usepackage{algpseudocode}%
\usepackage{listings}%
\tcbset{
  mycssstyle/.style={
    colback=black!5!white,
    colframe=black!75!black,
    listing only,
    listing options={%
      language=HTML,
      basicstyle=\ttfamily\small,
      breaklines=true
    },
    title=CSS Example
  }
}
\usepackage{graphicx}

\newcommand{\data}{\textsc{IndoTabVQA}}
\newcommand{\dataid}{\textsc{IndoTabVQA-id}}
\newcommand{\datae}{\textsc{IndoTabVQA-en}}
\newcommand{\datah}{\textsc{IndoTabVQA-hi}}
\newcommand{\dataar}{\textsc{IndoTabVQA-ar}}
\usepackage[T1]{fontenc}
\usepackage[utf8]{inputenc}

\usepackage{microtype}

\usepackage{inconsolata}

\usepackage{graphicx}

\title{\data{}: A Benchmark for Cross-Lingual Table Understanding in Bahasa Indonesia Documents}

\author{
 \textbf{Somraj Gautam\textsuperscript{1}},
 \textbf{Anathapindika Dravichi\textsuperscript{2}},
 \textbf{Gaurav Harit\textsuperscript{1}}
\\
 \textsuperscript{1}IIT Jodhpur,
 \textsuperscript{2}Punjabi University
\\
 \texttt{gautam.8@iitj.ac.in, dravichijan@gmail.com, gharit@iitj.ac.in}\\
  \href{https://huggingface.co/datasets/NusaBharat/INDOTABVQA}{\textbf{{https://huggingface.co/datasets/NusaBharat/INDOTABVQA}}
 }
}

\begin{document}
\maketitle
\begin{abstract}
We introduce \data{}, a benchmark for evaluating cross-lingual Table Visual Question Answering (VQA) on real-world document images in Bahasa Indonesia. The dataset comprises 1,593 document images, 
% totaling 1,910 tables 
across three visual styles (bordered, borderless, and colorful) with one or more than one tables, and 1,593 question-answer sets in four languages: Bahasa Indonesia, English, Hindi, and Arabic. This enables evaluation of Vision-Language Models (VLMs) in both monolingual (Bahasa documents with Bahasa questions) and cross-lingual settings (Bahasa documents with questions in other languages). We benchmark leading open-source VLMs (Qwen2.5-VL, Gemma-3, LLaMA-3.2) and GPT-4o and reveal substantial performance gaps, particularly on structurally complex tables and in low-resource languages. Fine-tuning a compact 3B and LoRA-finetuned 7B model on our dataset yields 11.6\% and 17.8\% improvements in accuracy. Providing explicit table region coordinates as additional input further improves performance by 4-7\%, demonstrating the value of Spatial priors for table-based reasoning. Our findings underscore the importance of language-diverse, domain-specific datasets and demonstrate that targeted fine-tuning can significantly enhance VLM performance on specialized document understanding tasks. \data{} provides a valuable resource for advancing research in cross-lingual, structure-aware document understanding, especially in underrepresented regions of the world. Full dataset can be accessed at:~\textcolor{blue}{\url{https://huggingface.co/datasets/NusaBharat/INDOTABVQA}}.
\end{abstract}

\section{Introduction}
\begin{figure}[t]
    \centering
    \includegraphics[width=0.44\textwidth]{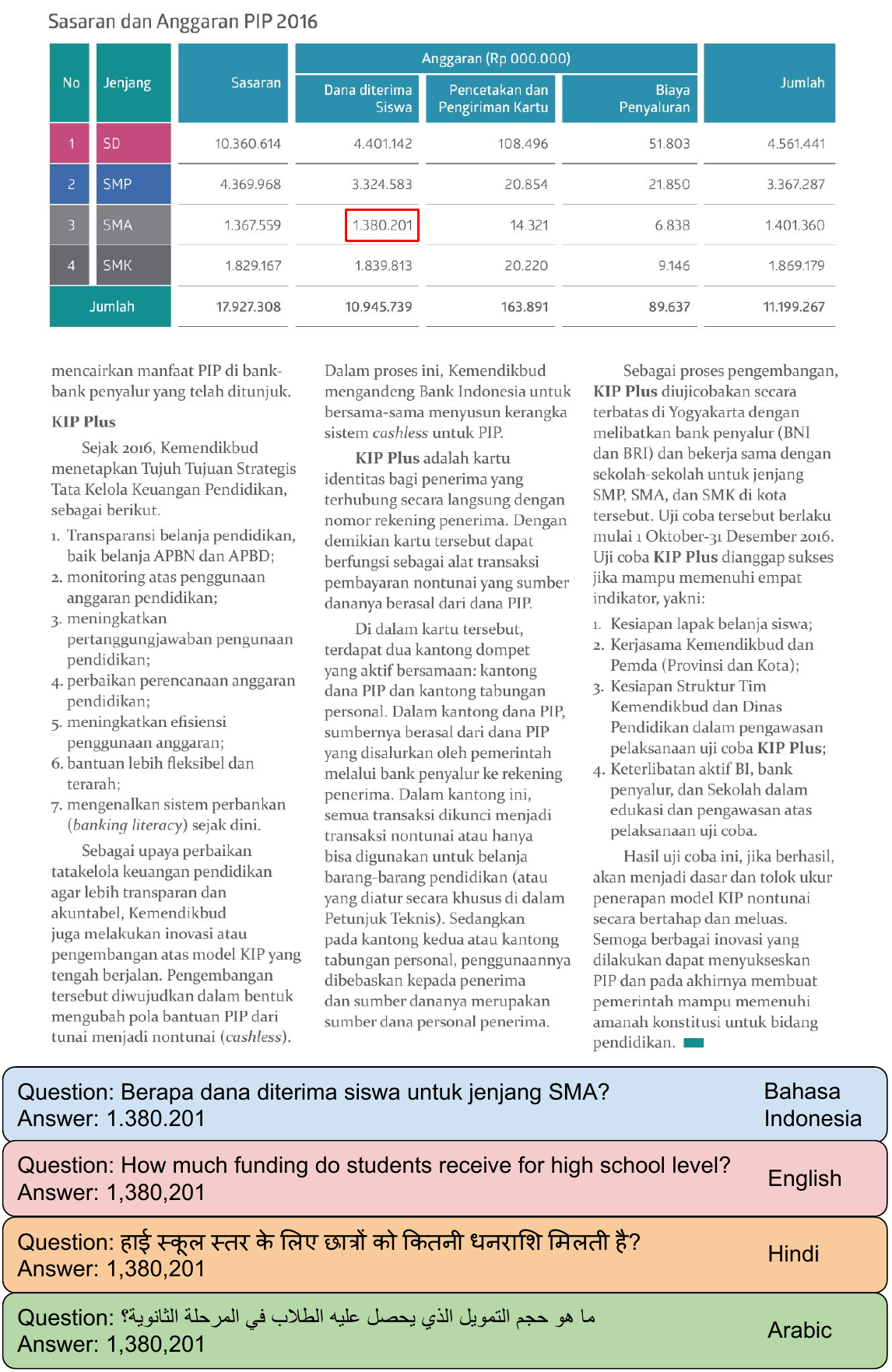}
    \caption{\data{} presents document images in Bahasa Indonesia, and semantically aligned QA pairs in four languages, enabling cross-lingual evaluation of VLMs.}
    \label{enter-label}
\end{figure}
Vision-Language Models (VLMs) have demonstrated strong performance on text-centric visual understanding tasks, as shown on benchmarks such as TextVQA~\cite{singh2019towards}, ST-VQA~\cite{xia2023st}, DocVQA~\cite{mathew2021docvqa}, and OCRBench~\cite{liu2024ocrbench}. Recent table-focused datasets such as TableVQA-Bench~\cite{kim2024tablevqa}, TabComp~\cite{gautam2025tabcomp}, and ComTQA~\cite{zhao2024tabpedia} further assess numerical reasoning and structure-aware comprehension. However, these benchmarks share a critical limitation: they are predominantly monolingual and English-centric, providing limited insight into VLM performance on low-resource languages or cross-lingual generalization. %This limitation has significant real-world implications. 
Documents in languages like Bahasa Indonesia, Hindi, and Arabic represent billions of users globally, yet VLMs trained primarily on English data may fail to process these documents reliably. For table-based VQA specifically, models must handle both linguistic variation and structural complexity, a challenging combination that remains underexplored.

\textbf{The Core Problem:} Existing VQA benchmarks do not adequately test whether VLMs can (1) understand tables in low-resource languages, or (2) answer questions about these tables when queries are posed in different languages. This gap limits our understanding of true multilingual capability and hinders the development of globally applicable document AI systems.

This paper introduces \data{}, a benchmark designed to evaluate the cross-lingual and structure-aware capabilities of VLMs in the context of real-world document tables. Our benchmark comprises document images containing tables in Bahasa Indonesia, a language spoken by over 200 million people but underrepresented in vision-language research, paired with question-answer (QA) annotations in Bahasa Indonesia, English, Hindi, and Arabic, as shown in Fig.~\ref{enter-label}. Detailed statistics of our benchmark are presented in section~\ref{Data_stat} and table~\ref{tab:dataset-comparison}.

% Our dataset comprises 1,593 real-world document images containing tables in Bahasa Indonesia, sourced from government reports, business documents, and educational records, with cross-lingual question-answer pairs in four languages (Bahasa Indonesia, English, Hindi, and Arabic). The dataset provides visual diversity through bordered, borderless, and colorful tables reflecting real-world document variation, along with both monolingual and cross-lingual evaluation settings to separately assess visual understanding and language transfer capabilities.

% Through comprehensive evaluation, we observe that zero-shot performance remains limited, with state-of-the-art VLMs including GPT-4o showing substantial performance degradation on table-centric document images in Bahasa Indonesia (21-44\% accuracy across languages). However, task-specific fine-tuning on our dataset improves a 3B model's performance by 17.8 percentage points on average, demonstrating the value of domain-specific supervision for document understanding. We also find that providing explicit spatial information through table bounding box coordinates improves performance by 3-5\%, particularly on borderless tables where visual structure is ambiguous. Furthermore, table visual style significantly affects model performance, with models performing differently across bordered, borderless, and colorful tables, and borderless tables showing the largest performance variance.

Our work provides three main contributions:
\begin{itemize}
    \item \textbf{A novel cross-lingual benchmark} featuring real-world documents in an underrepresented language (Bahasa Indonesia) with parallel annotations in four languages, enabling systematic evaluation of cross-lingual visual reasoning.
    \item \textbf{Comprehensive baseline evaluation} of current VLMs, revealing specific failure modes in structure-aware reasoning and language transfer that inform future model development.
    \item \textbf{Analysis of Spatial priors and fine-tuning} showing that explicit table localization and domain adaptation are effective strategies for improving VLM performance on specialized document tasks.
\end{itemize}
\data{} addresses a critical gap in multilingual document AI and provides a testbed for developing more inclusive and robust vision-language systems. The dataset and evaluation code will be made publicly available upon acceptance.

\section{\data{} Dataset}
This section describes the construction of \data{} in detail, covering the dataset scope and design, data collection, the diversity of table types, the annotation protocol, statistics, and benchmark configuration. 
% Table~\ref{tab:dataset-comparison} summarizes the key dataset properties.

\subsection{Dataset Scope and Design}
\data{} enables evaluation in two settings:
\begin{itemize}
    \item \textbf{Monolingual setting:} Both documents and QA pairs are in Bahasa Indonesia, testing the model's ability to understand low-resource language content.
    \item \textbf{Cross-lingual setting:} Documents remain in Bahasa Indonesia while questions are posed in English, Hindi, or Arabic. This probes whether models can align visual content in one language with semantically equivalent questions in another, assessing true cross-lingual transfer rather than memorized language patterns.
\end{itemize}

This design isolates two distinct challenges: (1) visual-linguistic understanding of low-resource document content, and (2) cross-lingual alignment between visual and textual modalities.

\subsection{Data Collection and Sources}
We sourced table images from real-world Indonesian documents across government reports (statistical summaries, budget allocations), educational records (enrollment data, performance metrics), business documents (invoices, financial statements), public health data (demographic statistics, service records). A significant portion of our data derives from the Institutional Repository of the Ministry of Primary and Secondary Education of Indonesia\footnote{\url{https://repositori.kemendikdasmen.go.id}}. We retrieved documents from the official portal and manually selected those containing well-formed tables suitable for VQA. %This process yielded 1,593 high-quality document images containing 1,910 total tables (average 1.20 tables per image).

% All documents are real-world materials in Bahasa Indonesia, often featuring local terminology and abbreviations, with quality ranging from clean printed forms to scanned documents with minor artifacts. Images are stored in PNG format at OCR-compatible resolution (300 DPI minimum).

\begin{figure*}
\centering
\includegraphics[width=0.85\textwidth]{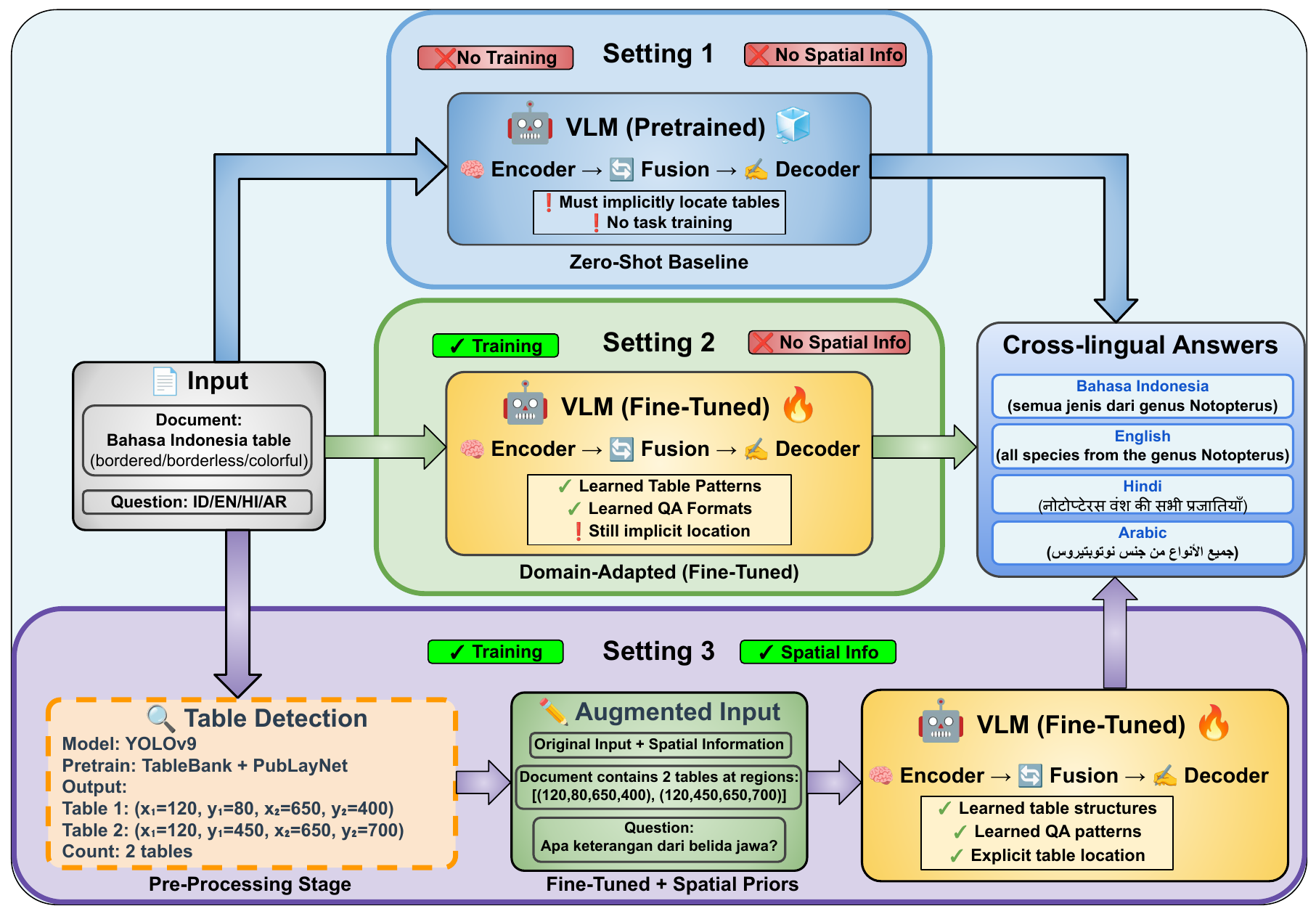}
\caption{Architecture comparison with left-to-right pipeline flow across three evaluation settings. Each row represents a complete evaluation pipeline from input to output. \textbf{Setting 1:} Uses a pretrained VLM without fine-tuning (zero-shot). \textbf{Setting 2:} Fine-tunes the model on table QA data but without spatial information. \textbf{Setting 3:} Introduces spatial priors through table detection, enabling the model to use table locations during reasoning.}
\label{main_diag}
\end{figure*}

\subsection{Visual Diversity: Table Types}\label{tabletype}
To reflect real-world document variation, we categorize tables into three types based on visual presentation:

\begin{itemize}
    \item \textbf{Bordered Tables} (500 images): Traditional tables with explicit cell borders, commonly found in official forms and reports. %Clear visual structure makes row-column relationships explicit.
    \item \textbf{Borderless Tables} (602 images): Tables without explicit cell lines, requiring inference of structure from whitespace, alignment, and text positioning. %These are common in modern documents and reports but pose significant challenges for visual understanding.
    \item \textbf{Colorful Tables} (491 images): Tables using background colors, cell shading, or highlighted headers for emphasis or grouping. %Color serves as additional structural information but may interfere with text recognition.
\end{itemize}
This taxonomy is not mutually exclusive (some tables have both borderless and colors as table present in Fig.~\ref{enter-label}), but we assign each image to its primary category for analysis purposes and give the color category a higher priority.

\subsection{Annotation Protocol} \label{ANN}
Each table instance is paired with one question–answer (QA) item, authored in Bahasa Indonesia, following a controlled template designed to cover lookup, aggregation, comparison, and structural reasoning. Annotators were instructed to write unambiguous, table-grounded, and answer-contained questions. We then translated each Bahasa QA into English, Hindi, and Arabic using automatic translation, followed by human validation by native speakers. Validators corrected lexical errors, normalized number formats, ensured that entity references remained faithful to the table, and flagged ambiguous or culturally mismatched translations. Each QA underwent a two-stage quality check: (1) internal consistency (answer must exist exactly in the table region) and (2) cross-lingual equivalence (the four versions must express the same intent). Items failing either check were revised or removed. Table~\ref{annotation} summarises the key statistics. Extended guidelines and annotation examples appear in Appendix~\ref{sec:appendix}. Figure~\ref{geo_chart} illustrates language coverage by country, highlighting our focus on evaluating VLMs in linguistically diverse and underrepresented regions, such as \textbf{Southeast Asia, the Middle East, South Asia, and other English-dominated countries}.

\subsection{Dataset Statistics and Properties}\label{Data_stat}
Table~\ref{tab:dataset-comparison} summarizes the key characteristics of the dataset. The visual content in all images is exclusively in Bahasa Indonesia, ensuring linguistic consistency across table elements. However, the question–answer (QA) annotations are multilingual, available in Bahasa Indonesia, English, Hindi, and Arabic, enabling cross-lingual evaluation and analysis. Each table instance is accompanied by detailed annotation metadata, including table-level bounding boxes to precisely locate tables within document images and table type tags covering three distinct categories that capture structural or functional variations among tables.

\begin{table}[t]
\scriptsize
\centering
\begin{tabular}{ll}
\toprule
\textbf{Property} & \textbf{\data{}} \\ \midrule
\# Document Images & 1,593 \\ 
\# Total Tables & 1,910 \\
Avg. Tables per Image & 1.20 \\
\# QA Pairs & 6,372 (Bahasa+English+Hindi+Arabic) \\
QA per Language & 1,593 per language \\
Languages & Bahasa Indonesia, English, Hindi, Arabic \\
QA Annotation Style & Human-written + Translated \\
Table Layouts & Bordered, Borderless, Colorful table \\
Domains & Government, Finance, Education, Health \\
Image Format & JPEG (OCR-compatible resolution) \\
Bounding Box Annotations & Table-level bounding boxes \\
Cross-lingual Setting & Doc in Bahasa, QA in other languages \\
\bottomrule
\end{tabular}
\caption{\data{} dataset properties covering multilingual QA, layout styles, and domain diversity.}
\label{tab:dataset-comparison}
\end{table}

\subsection{Benchmark Configuration}
We split the dataset into Test/Training/Validation set: 1043/500/50 samples.
    
We intentionally maintain a large test set to enable robust evaluation across diverse table styles and domains. Also, a small training set proves that fine-tuning with a small dataset size can improve the model's capability effectively.

\section{Evaluation Methodology}
\subsection{Task Formulation}
We formulate the task as image-grounded visual question answering: Given a document image $I$ containing one or more tables and a natural language question $Q$, in language $L \in \{\text{Bahasa Indonesia}, \text{English}, \text{Hindi}, \text{Arabic}\}$, the model must generate or select the correct answer $A$ in the same language. Formally, the task can be described as:
\[
A = \text{VLM}(I, Q)
\]
\subsection{Input Format}
Each input instance consists of a table image $I$ (in PNG or JPEG format) and a question $Q$ in either Bahasa Indonesia, English, Hindi, or Arabic. The answer A is a short free-form text or numeric value. Question types span factual lookup (retrieving specific cell values), numerical comparison (identifying maximum, minimum, or ranking items), aggregation (sum, count, or computing over multiple cells), and table-structure-related queries about table organization or headers.

\subsection{Evaluation Settings}\label{eval}
We evaluate models under three settings shown in Fig.~\ref{main_diag}:

\begin{itemize}
    \item \textbf{Zero-Shot Evaluation:} Models are tested directly on \data{} without any task-specific training. This measures out-of-the-box capability for cross-lingual table understanding.
    
    \item \textbf{Fine-Tuned Evaluation:} Model is trained on the \data{} training set (500 images) and evaluated on the test set (1,043 images).
    
    \item \textbf{Fine-Tuned + Spatial Priors:} In this, we add an explicit table detection pre-processing stage (orange block) using YOLOv9 to locate table regions. These coordinates are then incorporated into an augmented prompt before VLM processing. 
\end{itemize}

\subsection{Table Localization as Additional Input}
\textbf{Motivation:} A key challenge in document VQA is that tables may occupy only a small region of the full image, and documents may contain multiple tables with varying layouts and positions. Real-world document processing systems typically address this through multi-stage pipelines that first detect document regions (tables, figures, text blocks) before applying specialized models to each region. By providing explicit table bounding box coordinates to VLMs, we mirror this practical workflow and potentially help models focus their attention on relevant content rather than searching across the entire image. 
This approach also allows us to isolate the impact of spatial localization from other factors affecting model performance, providing insights into whether structural ambiguity, particularly in borderless tables, is a primary bottleneck for accurate table understanding.

\subsection{Implementation of setting 3}
Our approach consists of two stages:
\begin{itemize}
    \item \textbf{Stage 1: Table Detection:} A separate, off-the-shelf object detection model (YOLOv9~\cite{wang2024yolov9} pretrained on TableBank~\cite{li2020tablebank} and PubLayNet~\cite{zhong2019publaynet}) to identify table regions in document images. The detector outputs: 1) Bounding box coordinates: [$(x_1, y_1, x_2, y_2)$, ...] for each detected table. 2) Number of tables detected: N
    
    \item \textbf{Stage 2: Augmented Input:} VLM receives:1) Original input + Table bounding boxes. 2) Number of tables.
\end{itemize}
Example prompt augmentation:
\begin{tcolorbox}[mycssstyle, title=Augmented Input]
Document Image $I$.

Document contains 2 tables at regions: [(120, 80, 650, 400), (120, 450, 650, 700)].

Question: [question text]

Answer:
\end{tcolorbox}

\subsection{Evaluation Metrics}

To evaluate model performance across diverse settings and languages, we employ both exact and semantic answer matching strategies.

\subsubsection{In-Match Accuracy (Relaxed Matching)}  
We use a relaxed matching criterion where a prediction is correct if the normalized ground truth answer appears as a substring within the predicted answer.

Normalization involves converting text to lowercase, removing punctuation, collapsing whitespace, and handling number formatting variations. This relaxed matching accounts for VLMs that often generate answers with additional context (e.g., `if the ground truth is `5 tables', a prediction of `There are 5 tables in the document') would be considered correct. In-Match captures correct answers embedded in longer responses.

\noindent\textbf{Formula:}

{\small
\begin{equation}
\text{In-Match}(A_p, A_g)=
\begin{cases}
1, & \text{Norm}(A_g)\subseteq \text{Norm}(A_p),\\
0, & \text{otherwise}.
\end{cases}
\end{equation}
}
\begin{table*}[!t]
\scriptsize
    \centering
    \begin{tabular}{l c c c c >{\columncolor[gray]{0.9}}c c c c c >{\columncolor[gray]{0.9}}c}
    \toprule
    & \multicolumn{5}{c}{\textbf{In-Match accuracy(\%)}} 
    & \multicolumn{5}{c}{\textbf{STS Accuracy(\%)}} \\
    \cmidrule(lr){2-6} \cmidrule(lr){7-11}
    \textbf{Model [\#params]} 
    & ID & EN & HI & AR & $\Delta$
    & ID & EN & HI & AR & $\Delta$ \\
    \midrule
    \multicolumn{11}{c}{\cellcolor[gray]{0.9}{\textbf{Open-source}}} \\
    \midrule
    Donut
      & 10.5 & 5.48 & 4.74 & 4.39 & 6.20 
      & 15.52 & 9.10 & 5.17 & 6.03 & 8.96 \\
    Qwen2.5VL [3B] 
      & 37.8 & 28.7 & 4.1 & 16.4 & 21.9 
      & 29.0 & 44.9 & 4.4 & 27.5 & 26.5 \\
    Gemma3 [12B] 
      & 40.9  & 27.4  & 19.5  & 17.4  &  26.1
      & 41.4 & 31.0  & 27.3 & 26.5 & 31.6 \\
    Qwen2.5VL [7B] 
      & 54.8  & 36.2  & 17.3  & 23.0  & 32.9  
      & 36.5 & 58.1  & 16.1  & 34.3 & 36.3 \\
    Llama-3.2 [11B] 
      & 57.4  & 30.8  & 15.5  & 19.4  & 30.7 
      & 54.2 & 36.1 & 15.7  & 19.5 & 31.4 \\ \midrule
      \multicolumn{11}{c}{\cellcolor[gray]{0.9}{\textbf{Closed-source}}} \\ \midrule
    GPT-4o
      & 72.2  & 44.6  & 26.0  & 21.4  & 41.1
      & 71.1 & 60.6 & 38.8  & 38.4 & 52.2 \\      \midrule
      \multicolumn{11}{c}{\cellcolor[gray]{0.9}{\textbf{Finetuned + Spatial Priors (SP)}}} \\
      \midrule
    GPT-4o+SP
      & 72.6  & 52.7  & \underline{27.2}  & 25.5  & 44.6
      & 73.4 & 62.2  & 39.1 & 40.0 & 53.6 \\
    \data{} [3B]
      & 66.4  & 46.1   & 22.1  & 25.8  &  39.7
      & 71.4 & 49.3 & 27.3  & 38.0 & 46.7 \\
    \data{} [7B] 
      & 71.9  & 51.6   & 26.2  & 28.1  & 44.5 & \underline{77.6} & \underline{64.5} & 31.4 & \underline{46.4} & \underline{54.9} \\
    \data{} [3B]+SP
      & \underline{73.1}  & \underline{54.8} & \underline{27.2}  & \underline{31.1}  & \underline{46.6} & 75.2 & 61.2  & \underline{36.0} & 40.1 & 53.1 \\
    \data{} [7B]+SP 
      & \textbf{78.3}  & \textbf{58.4} & \textbf{29.4}  & \textbf{32.8}  & \textbf{48.5} & \textbf{82.1} & \textbf{66.1} & \textbf{36.7} & \textbf{48.6} & \textbf{58.3} \\
    \bottomrule
    \end{tabular}
    \caption{\label{tab:result1} Evaluation of various VLMs on In-Match and STS Accuracy across four languages here ID is \dataid, EN is \datae, HI is \datah, AR is \dataar, and SP is Spatial Prior, $\Delta$ is average accuracy.}
\end{table*}
\subsubsection{Semantic Textual Similarity (STS)}  
To better assess how well our model captures the true meaning of an answer, we go beyond simple word-for-word comparisons. We use \textbf{Semantic Textual Similarity (STS)} to measure the degree of meaning alignment between predicted answers \( A_p \) and ground truth answers \( A_g \). STS is computed as the cosine similarity between their dense vector representations:
\begin{equation}
\text{STS}(A_p, A_g) = \frac{ \phi(A_p) \cdot \phi(A_g) }{ \|\phi(A_p)\| \cdot \|\phi(A_g)\| }
\end{equation}
where \( \phi(\cdot) \) denotes a sentence-level semantic encoder. To compute Semantic Textual Similarity (STS), we use the \texttt{paraphrase-multilingual-MiniLM-L12-v2} model from the \textbf{Sentence Transformers} library \cite{reimers2019sentencebert}, which produces language-agnostic sentence embeddings across 50+ languages. The similarity score lies in \([0, 1]\), with higher values indicating greater semantic alignment.

\paragraph{Breakdown by QA Type.}  
Beyond overall metrics, to understand where models succeed and fail, we report fine-grained accuracy across all four languages, table types mentioned in section~\ref{tabletype}, and evaluation Settings mentioned in section~\ref{eval}
\subsection{Baseline Models}
We evaluate a diverse set of VLMs spanning different scales and architectures, including Open-Source Models such as:
\textbf{Qwen2.5-VL [3B]}~\cite{wang2024qwen2}: Compact VLM with strong multilingual capability. \textbf{Qwen2.5-VL [7B]}~\cite{bai2025qwen2}: Larger variant with enhanced reasoning. \textbf{Gemma-3 [12B]}~\cite{team2025gemma}: Google's model with broad language coverage. \textbf{LLaMA-3.2 [11B]}~\cite{grattafiori2024llama}: Meta's vision-enabled language model, and a Closed-Source Model such as
GPT-4o~\cite{openai2024gpt4api}, which is a state-of-the-art proprietary VLM with strong multilingual performance.

% This suite covers a Scale range of 3B to 12B+ parameters, including different architectures Such as Encoder-Decoder and decoder-only models, both open (reproducible) and closed (performance ceiling) models, and models with demonstrated multilingual capabilities.

We also evaluate Donut (Document Understanding Transformer~\cite{ocr_free}), an OCR-free document understanding model that directly maps document images to structured outputs using an encoder-decoder architecture. As it lacks multilingual pretraining and cross-lingual transfer capabilities, we expect it to serve as a lower-bound baseline, particularly in the Hindi and Arabic settings.

\subsection{Fine-Tuning Configuration}
% We fine-tune Qwen2.5-VL [3B] as our primary model due to 
% manageable computational requirements (trainable on a single GPU), open weights and architecture (reproducible), strong baseline performance, and multilingual pretraining. 
Our fine-tuning strategy follows a full Instruction Fine-Tuning approach for Qwen2.5-VL [3B] and parameter-efficient finetuning (LoRA) for Qwen2.5-VL [7B]. The model was trained separately on each language variant of the dataset to isolate language-specific learning patterns. A detailed training setup is present in Appendix~\ref{Train_setup}

\section{Results and Analysis}
We present evaluation results across three dimensions: (1) overall performance by language, (2) breakdown by table visual style, and (3) fine-grained analysis by question type. Our analysis focuses on understanding where and why models struggle, rather than simply ranking performance. Results are reported using two complementary metrics: In-Match accuracy, which measures relaxed answer inclusion, and STS accuracy, which captures semantic similarity using sentence-level embeddings. Our analysis spans both language-wise performance (Table~\ref{tab:result1}) and table-type-specific behavior across languages (Table~\ref{tab:multiimage_res}).

\subsection{Overall Performance Across Languages}
Table~\ref{tab:result1} presents In-Match and STS accuracy for all models across four languages. Several patterns emerge:

\subsubsection{Performance Ranking by Model Scale (Zero-Shot)}
Zero-shot performance among open-source VLMs generally increases with model scale. 
\textbf{Qwen2.5-VL-3B} attains 21.9\% average accuracy, while \textbf{Qwen2.5-VL-7B} improves to 32.9\%. 
Larger models such as \textbf{LLaMA-3.2-11B} and \textbf{Gemma-3-12B} achieve intermediate performance (26--31\%).

However, scale alone is insufficient: \textbf{Qwen2.5-VL-7B} outperforms the larger \textbf{Gemma-3-12B}, underscoring the importance of architecture and pretraining. 
\textbf{GPT-4o} delivers the best zero-shot results (41.1\% In-Match, 52.2\% STS), reflecting the benefits of large-scale, diverse training.

\begin{table*}[!t]
\scriptsize
\centering
\begin{tabular}{l c c c !{\vrule width 1pt} c c c}
\toprule
 & \multicolumn{3}{c|}{\textbf{In-Match accuracy(\%)}} & \multicolumn{3}{c}{\textbf{STS Accuracy(\%)}} \\ \cmidrule(lr){2-4} \cmidrule(lr){5-7} 
\textbf{Model [\#params]} & Bordered & Borderless & Colorful & Bordered & Borderless & Colorful \\ \midrule
\multicolumn{7}{c}{\cellcolor[gray]{0.9}{\textbf{Bahasa Indonesia}}} \\ \midrule
Donut & 11.71 & 10.23 & 9.40 & 21.02  & 17.87 & 8.95 \\
Qwen2.5VL-3B & 32.73 & 44.25 & 36.36 & 24.62 & 34.27 & 27.9 \\
Qwen2.5VL-7B & 52.55 & 57.29 & 54.55 & 40.54 & 32.74 & 36.36 \\
Gemma3-12B & 48.05 & 34.78 & 39.81 & 48.35 & 32.74 & 43.26 \\
Llama-3.2-11B & 57.36 & 52.43 & 62.38 & 52.20 & 50.50 & 60.10 \\
GPT-4o & 74.03 & 65.94 & 76.60 & 71.47 & 70.08 & 71.16 \\ 
\rowcolor{blue!10}
\multicolumn{7}{c}{\textbf{Finetuned + Spatial Priors (SP)}}  \\ 
\textbf{GPT-4o}+SP & 75.23 & 65.94 & 76.6 & 73.38 & 73.45 & 73.34 \\ 
\textbf{\dataid}~[3B] & 72.07 & 62.92 & 64.26 & 73.38 & 68.43 & 72.27 \\
\textbf{\dataid}~[7B] & 72.07 & 69.82 & 74.61 & 79.58 & 73.15 & 81.50 \\
\textbf{\dataid}~[3B]+SP & 80.78 & 66.75 & 71.79 & 81.38 & 71.59 & 72.73 \\
\textbf{\dataid}~[7B]+SP & 80.25 & 73.15 & 81.50 & 87.65 & 76.47 & 83.39 \\
\midrule

\multicolumn{7}{c}{\cellcolor[gray]{0.9}{\textbf{English}}} \\ \midrule
Donut & 3.90 & 5.63 & 6.90 & 7.81 & 6.14 & 14.11 \\
Qwen2.5VL-3B & 20.72 & 33.76 & 31.66 & 44.74 & 45.27 & 44.51 \\
Qwen2.5VL-7B & 29.43 & 41.18 & 37.93 & 55.26 & 61.13 & 58 \\
Gemma3-12B & 27.03 & 23.53 & 31.66 & 32.43 & 26.60 & 33.86 \\
Llama-3.2-11B & 25.53 & 28.90 & 37.93 & 41.40 & 28.90 & 37.93 \\
GPT-4o & 42.34 & 41.18 & 50.16 & 63.96 & 53.96 & 63.95 \\ 
\rowcolor{blue!10}
\multicolumn{7}{c}{\textbf{Finetuned + Spatial Priors (SP)}}  \\ 
\textbf{GPT-4o}+SP & 42.81 & 56.42 & 58.87 & 65.23 & 55.12 & 66.17 \\
\textbf{\datae}~[3B] & 37.84 & 54.73 & 45.77 & 50.15 & 47.31 & 50.47 \\
\textbf{\datae}~[7B] & 45.35 & 52.45 & 56.87 & 63.06 & 65.73 & 64.58 \\
\textbf{\datae}~[3B]+SP & 48.95 & 55.75 & 59.87 & 60.70 & 57.30 & 65.50 \\
\textbf{\datae}~[7B]+SP & 53.85 & 58.75 & 62.57 & 64.86 & 66.75 & 66.77 \\
\midrule

\multicolumn{7}{c}{\cellcolor[gray]{0.9}{\textbf{Hindi}}} \\ \midrule
Donut & 3.60 & 4.35 & 6.27 & 4.20 & 4.60 & 6.90 \\
Qwen2.5VL-3B & 3.90 & 4.60 & 3.45 & 2.70 & 6.90 & 3.50 \\
Qwen2.5VL-7B & 14.41 & 18.41 & 18.81 & 13.81 & 16.88 & 17.55 \\
Gemma3-12B & 16.50 & 17.40 & 24.50 & 26.43 & 23.53 & 32.29 \\
Llama-3.2-11B & 12.91 & 13.04 & 21.32 & 12.91 & 13.04 & 21.32 \\
GPT-4o & 20.92 & 26.80 & 30.35 & 35.44 & 39.62 & 40.22 \\ 
\rowcolor{blue!10}
\multicolumn{7}{c}{\textbf{Finetuned + Spatial Priors (SP)}}  \\ 
\textbf{GPT-4o}+SP & 22.32 & 28.52 & 30.76 & 36.9 & 37.39 & 43.01 \\
\textbf{\datah}~[3B] & 13.21 & 25.58 & 20.38 & 18.92 & 35.04 & 27.59 \\
\textbf{\datah}~[7B] & 20.42 & 29.92 & 28.21 & 25.53 & 33.76 & 34.80 \\
\textbf{\datah}~[3B]+SP & 14.11 & 28.90 & 21.94 & 31.02 & 36.88 & 40.1 \\
\textbf{\datah}~[7B]+SP & 22.82 & 33.76 & 31.66 & 27.93 & 41.18 & 40.44 \\
\midrule

\multicolumn{7}{c}{\cellcolor[gray]{0.9}{\textbf{Arabic}}} \\ \midrule
Donut & 2.10 & 5.96 & 5.12 & 6.61 & 3.84 & 8.15 \\
Qwen2.5VL-3B & 12.01 & 18.93 & 17.87 & 24.30 & 34.30 & 23.80 \\
Qwen2.5VL-7B & 19.50 & 24.81 & 24.14 & 28.23 & 41.43 & 33.23 \\
Gemma3-12B & 18.30 & 14.60 & 19.40 & 26.40 & 24.80 & 28.20 \\
Llama-3.2-11B & 15.92 & 18.16 & 24.45 & 15.92 & 18.16 & 24.45 \\
GPT-4o & 18.92 & 21.48 & 23.82 & 35.44 & 39.62 & 40.22 \\
\rowcolor{blue!10}
\multicolumn{7}{c}{\textbf{Finetuned + Spatial Priors (SP)}}  \\
\textbf{GPT-4o}+SP & 21.24 & 28.80 & 26.40 & 37.60 & 38.39 & 44.01 \\
\textbf{\dataar}~[3B] & 17.72 & 34.02 & 24.14 & 32.10 & 46.30 & 35.40 \\
\textbf{\dataar}~[7B] & 40.66 & 34.17 & 23.42 & 43.84 & 48.85 & 46.08 \\
\textbf{\dataar}~[3B]+SP & 21.32 & 39.64 & 32.29 & 35.44 & 45.78 & 39.18 \\
\textbf{\dataar}~[7B]+SP & 40.66 & 34.17 & 23.42 & 47.15 & 51.66 & 46.39 \\
\bottomrule
\end{tabular}
\caption{\label{tab:multiimage_res} Results of various VLMs on In-Match and STS Accuracy based on table types across four languages.}
\end{table*}
\subsection{The Cross-Lingual Performance Gap}\label{sec:4.2}
Performance drops substantially in cross-lingual settings compared to monolingual (Bahasa):
\subsubsection{Zero-shot degradation from ID to other languages:}
GPT-4o: 72.2\% $\rightarrow$ 44.6\% (EN), 26.0\% (HI), 21.4\% (AR), Qwen2.5-VL [7B]: 54.8\% $\rightarrow$ 36.2\% (EN), 17.3\% (HI), 23.0\% (AR)

This 30-50 percentage point drop reveals a critical limitation: models struggle to align visual content in one language with questions in another.
This led to two research questions (RQ) mentioned below:

\noindent \textbf{RQ1: Why is Hindi particularly difficult?}

Hindi shows the lowest accuracy across nearly all models (4-27.2\%). Possible explanations can \textbf{Script unfamiliarity:} Devanagari script is less common in VLM pretraining, and most mainstream models use subword tokenization algorithms like SentencePiece or BPE. When applied to Devanagari, these tokenizers often fail to identify meaningful morphological units, instead splitting words into a long sequence of less meaningful, sometimes single-character, tokens. This sub-optimal segmentation has two detrimental effects: first, it creates much longer input sequences for the model, increasing computational load and making it harder to capture long-range dependencies; second, and more importantly, it fails to provide the model with consistent, semantically meaningful representations for Hindi words and concepts, thereby hindering learning and generalization~\cite{kanjirangat2025tokenization}.

Similarly, the challenges with Arabic extend beyond simple script differences. The Arabic language is a right-to-left (RTL) script, which can confound models that implicitly assume a left-to-right flow of information, especially for questions involving spatial relationships.

\noindent \textbf{RQ2: Why does Bahasa perform best?}

The monolingual setting removes the cross-lingual alignment challenge, as both the visual content and the question share the same language. Additionally, the fine-tuned model is directly exposed to Bahasa examples during training, giving it a distributional advantage. As shown in Table 2, language-specific In-Match gains after fine-tuning are: Bahasa Indonesia +28.6 points (highest), English +17.4, Hindi +18.0, and Arabic +9.4, demonstrating that even modest task-specific supervision over 500 training images yields meaningful improvements across all languages.

\subsection{Effect of Spatial Priors (Bounding Boxes)}
Adding table bounding box coordinates as additional input provides further gains:

\textbf{Average improvement over fine-tuned model:}
Compact 3B Model: In-Match: +6.9\% points (39.7\% $\rightarrow$ 46.6\%), 
STS: +6.4\% points (46.7\% $\rightarrow$ 53.1\%).
LoRA Finetuned 7B Model: In-Match: +4.0\% points (44.5\% $\rightarrow$ 48.5\%), STS: +3.4\% points (54.9\% $\rightarrow$ 58.3\%).

\textbf{Average improvement on GPT-4o model:}
In-Match: +3.5\% points (41.1\% $\rightarrow$ 44.6\%), 
STS: +1.4\% points (52.2\% $\rightarrow$ 53.6\%)

Notably, spatial priors benefit GPT-4o as well, boosting its In-Match accuracy from 41.1\% to 44.6\%, confirming that explicit table localization is useful regardless of model scale. Our fine-tuned 7B model with spatial priors achieves the best overall performance across both metrics (48.5\% In-Match and 58.3\% STS), outperforming GPT-4o+SP (44.6\% and 53.6\% respectively), suggesting that the combination of domain adaptation and spatial grounding is more effective than either alone.

\subsection{Performance by Table Visual Style}
Table~\ref{tab:multiimage_res} analyzes model performance across bordered, borderless, and colorful tables, revealing the strong influence of visual style on reasoning accuracy. Borderless tables pose the greatest challenge, as models must infer row–column relationships from whitespace and alignment, often leading to ambiguity. Accuracy improves notably with spatial priors (e.g., +3.8 points in Bahasa), showing the benefit of explicit localization. Colorful tables yield mixed results; GPT-4o performs better on them (76.6\% vs. 74.0\%), likely because color aids visual grouping and attention, though smaller models struggle due to limited robustness to color variation. Bordered tables provide the clearest structure and serve as a baseline (GPT-4o: 74.0\%, LLaMA-3.2: 57.4\%, Gemma3-12B: 48.1\%, Qwen2.5-VL [7B]: 52.6\%). Yet even here, performance below 75\% indicates that accurate table reasoning remains a challenging task despite clear visual cues. As shown in Fig.~\ref{last_diag}, our model+SP produces correct results across all three types of tables in a cross-lingual setting.

To better understand model failures in our benchmark, we perform a detailed manual analysis of erroneous predictions in section~\ref{error} of Appendix~\ref{sec:appendix}. We categorize the errors into five types for English, Hindi, and Arabic, and into four types for Bahasa Indonesia (due to the absence of translation-related errors) as shown in Figure~\ref{pie_chart} and Table~\ref{tab:error_table} in Appendix~\ref{sec:appendix}.

\begin{table*}[!t]
\scriptsize
\centering

\begin{minipage}{0.48\textwidth}
\centering
\begin{tabular}{l c c c c}
\toprule
\textbf{Model [\#params]} & Agg. & Comp. & Look. & Str. \\ 
\midrule

\multicolumn{5}{c}{\cellcolor[gray]{0.9}{\textbf{Indonesia}}} \\ 
\midrule
Donut & 13.13 & 0 & 4.89 & 12.33 \\
Qwen2.5VL-3B & 50.59 & 41.18 & 19.14 & 30.14 \\
Qwen2.5VL-7B & 64.58 & 76.47 & 27.56 & 57.53 \\
Gemma3-12B & 40.7 & 47.06 & 38.77 & 47.95 \\
Llama-3.2-11B & 66.23 & 100 & 43.25 & 53.42 \\
GPT-4o & 74.41 & 60.24 & 64.59 & 71.60 \\

\rowcolor{blue!10}
\multicolumn{5}{c}{\textbf{Finetuned + Spatial Priors (SP)}} \\

GPT-4o + SP & 75.48 & 64.71 & 78.89 & 72.60 \\
\dataid~[3B] & 71.92 & 47.06 & 58.97 & 65.75 \\
\dataid~[7B] & 76.62 & 82.35 & 65.55 & 71.23 \\
\dataid~[3B]+SP & 74.88 & 70.59 & 71.58 & 71.23 \\
\dataid~[7B]+SP & 80.39 & 70.59 & 78.54 & 82.19 \\

\midrule

\multicolumn{5}{c}{\cellcolor[gray]{0.9}{\textbf{English}}} \\ 
\midrule
Donut & 7.98 & 0 & 1.37 & 1.39 \\
Qwen2.5VL-3B & 40.96 & 41.18 & 11.85 & 12.50 \\
Qwen2.5VL-7B & 45.69 & 47.06 & 22.73 & 23.61 \\
Gemma3-12B & 32.39 & 23.53 & 20.43 & 20.83 \\
Llama-3.2-11B & 40.97 & 58.82 & 16.77 & 18.06 \\
GPT-4o & 54.72 & 41.18 & 35.97 & 37.50 \\

\rowcolor{blue!10}
\multicolumn{5}{c}{\textbf{Finetuned + Spatial Priors (SP)}} \\

GPT-4o + SP & 62.23 & 65.84 & 38.84 & 38.70 \\
\datae~[3B] & 58.26 & 41.18 & 26.83 & 31.94 \\
\datae~[7B] & 62.85 & 52.94 & 32.97 & 29.17 \\
\datae~[3B]+SP & 60.62 & 64.71 & 39.41 & 31.94 \\
\datae~[7B]+SP & 62.94 & 64.71 & 32.88 & 30.56 \\

\bottomrule
\end{tabular}
\end{minipage}
\hfill
\begin{minipage}{0.48\textwidth}
\centering
\begin{tabular}{l c c c c}
\toprule
\textbf{Model [\#params]} & Agg. & Comp. & Look. & Str. \\ 
\midrule

\multicolumn{5}{c}{\cellcolor[gray]{0.9}{\textbf{Hindi}}} \\ 
\midrule
Donut & 6.50 & 0 & 0.78 & 0 \\
Qwen2.5VL-3B & 6.40 & 0 & 2.56 & 3.36 \\
Qwen2.5VL-7B & 21.60 & 0 & 7.31 & 4.10 \\
Gemma3-12B & 21.26 & 29.41 & 11.78 & 6.85 \\
Llama-3.2-11B & 16.22 & 64.71 & 5.63 & 6.85 \\
GPT-4o & 32.22 & 29.41 & 14.87 & 9.59 \\

\rowcolor{blue!10}
\multicolumn{5}{c}{\textbf{Finetuned + Spatial Priors (SP)}} \\

GPT-4o + SP & 35.73 & 34.00 & 17.84 & 13.20 \\
\datah~[3B] & 29.01 & 29.41 & 5.40 & 5.48 \\
\datah~[7B] & 36.50 & 29.41 & 13.03 & 4.11 \\
\datah~[3B]+SP & 30.39 & 29.41 & 8.29 & 8.22 \\
\datah~[7B]+SP & 36.46 & 47.06 & 10.41 & 6.85 \\

\midrule

\multicolumn{5}{c}{\cellcolor[gray]{0.9}{\textbf{Arabic}}} \\ 
\midrule
Donut & 6.15 & 0 & 1.56 & 0 \\
Qwen2.5VL-3B & 21.50 & 5.88 & 5.28 & 4.11 \\
Qwen2.5VL-7B & 29.75 & 11.76 & 8.01 & 12.33 \\
Gemma3-12B & 20.72 & 35.29 & 7.42 & 12.33 \\
Llama-3.2-11B & 24.73 & 76.47 & 5.86 & 10.96 \\
GPT-4o & 23.75 & 47.06 & 10.74 & 17.81 \\

\rowcolor{blue!10}
\multicolumn{5}{c}{\textbf{Finetuned + Spatial Priors (SP)}} \\

GPT-4o + SP & 28.33 & 50.28 & 12.84 & 17.81 \\
\dataar~[3B] & 36.24 & 52.94 & 10.41 & 10.96 \\
\dataar~[7B] & 41.17 & 17.65 & 13.14 & 13.70 \\
\dataar~[3B]+SP & 38.62 & 35.29 & 11.34 & 9.59 \\
\dataar~[7B]+SP & 43.42 & 41.18 & 12.59 & 16.44 \\

\bottomrule
\end{tabular}
\end{minipage}

\caption{\label{question_type}Performance (In-match) across question types in four languages, Agg is Aggregation, Comp. is Comparison, Look. is Lookup, and Str. is Table structure related questions.}
\end{table*}

\subsection{Performance by Question Type}
We further analyze model behavior across four question types (lookup, aggregation, comparison, and structural reasoning) as shown in Table~\ref{question_type}. Consistent with earlier observations, aggregation and comparison questions achieve higher accuracy across models and languages, indicating that VLMs are relatively effective at coarse-grained reasoning where relevant values are localized or require limited structural interpretation. In contrast, lookup and structural questions remain more challenging. Lookup requires precise cell-level retrieval, while structural reasoning depends on understanding table organization, such as header alignment and row–column relationships. These challenges are further amplified in cross-lingual settings, where accurate alignment between the query language and table content is necessary. Hindi and Arabic exhibit the largest performance degradation, consistent with the cross-lingual trends discussed in Section~\ref{sec:4.2}. Fine-tuning improves performance across all question types, with the most notable gains in lookup and structural reasoning. Incorporating spatial priors provides additional improvements, particularly for lookup, by guiding the model toward relevant table regions. Overall, this analysis highlights that while current VLMs handle coarse reasoning well, fine-grained, structure-aware table understanding remains a key limitation.
% \subsection{Vision-Language Models and Multimodal Benchmarks}
% Recent years have seen rapid progress in vision-language models (VLMs), driven by the introduction of large-scale pretraining frameworks such as Flamingo~\cite{alayrac2022flamingo}, BLIP~\cite{li2022blip}, BLIP-2~\cite{li2023blip}, and PaLI~\cite{chen2023pali}. More recent efforts include InternVL~\cite{chen2024internvl}, LLaVA~\cite{liu2023visual}, MiniGPT-4~\cite{zhu2023minigpt}, and Qwen-VL~\cite{wang2024qwen2}, which extend vision-language capabilities to open-ended reasoning and zero-shot tasks. These models are typically evaluated on general-purpose benchmarks like VQAv2~\cite{goyal2017making}, OK-VQA~\cite{marino2019ok}, and GQA~\cite{hudson2019gqa}, which focus on natural images and mostly English-only question settings.
\begin{figure}[t]
    \centering
    \includegraphics[width=0.45\textwidth]{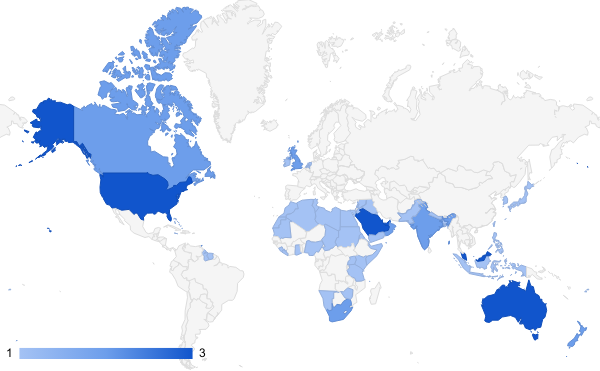}
    \caption{Global language coverage map for the \data{} benchmark. The shading intensity indicates the number of supported languages (1–3) spoken in each country. For example, Canada supports both English and Hindi. This visualization highlights the geographical and cultural reach of our cross-lingual benchmark.}
    \label{geo_chart}
\end{figure}
\section{Related Work}
\subsection{Table-Based Visual Question Answering}
Table-Based Visual Question Answering (VQA) addresses the challenge of reasoning over tabular structures embedded in images. Benchmarks such as InfographicVQA~\cite{mathew2022infographicvqa}, DocVQA~\cite{mathew2021docvqa}, and ChartQA~\cite{masry2022chartqa}, TabFact~\cite{chen2019tabfact}, TAT-QA~\cite{zhu2021tat}, and PubTables-1M~\cite{smock2022pubtables} emphasize reasoning over semi-structured and document tables. However, most existing benchmarks remain English-centric and fail to capture the visual noise, layout diversity, and multilingual characteristics of real-world documents.

\subsection{Multilingual and Cross-Lingual VQA}
Several benchmarks, including MULE~\cite{kim2020mule}, MTVQA~\cite{tang2025mtvqa} and MaXM~\cite{changpinyo2022maxm}, target multilingual captioning and VQA.
M\textsuperscript{4}C\allowbreak~\cite{kesen2025multilingual} considers multilingual documents but focuses primarily on scene text or scanned forms. %Recent work, such as xDoc~\cite{chen2022xdoc} and XFUND~\cite{xu2022xfund}, supports document tasks across several languages, though not specifically for table-based reasoning. 
Recent work on XT-VQA~\cite{yu2025cross} demonstrates the cross-lingual gap but is linguistically limited to Chinese, English, and French, whereas MMCricBench~\cite{gautam2025mind} is limited to English and Hindi only.
Both benchmarks share the goal of evaluating cross-lingual transfer, but \data{} provides complementary coverage of different languages, writing systems, and document types. Table~\ref{statistics2} contrasts \data{} with related benchmarks.

% \subsection{Datasets on Low-Resource Language}

% Despite increasing interest in inclusivity, most VQA and document understanding benchmarks remain heavily skewed toward English. Datasets like FUNSD~\cite{jaume2019funsd}, RVL-CDIP~\cite{harley2015evaluation}, and DocVQA~\cite{mathew2021docvqa} rarely include documents in languages like Bahasa Indonesia, Arabic, or Hindi. XFUND~\cite{xu2022xfund} and multilingual SQuAD~\cite{artetxe2019cross} are notable exceptions, but they lack document-level table images. Our benchmark fills this gap by introducing real-world document tables from underrepresented regions with cross-lingual QA supervision, providing a testbed for truly global VLM development.

\section{Conclusion}

We introduce \data{}, a novel benchmark for table-based VQA grounded in real-world document images from an underrepresented region, with cross-lingual QA pairs in Bahasa Indonesia, English, Hindi, and Arabic. Our evaluation reveals that even state-of-the-art closed-source VLMs like GPT-4o struggle with layout-aware and cross-lingual reasoning, particularly in low-resource languages. Fine-tuning a compact 3B model and LoRA-finetuning a 7B model on our dataset substantially improves performance. Our analysis shows that lookup and structural reasoning remain the hardest categories. The additional performance boost from spatial priors underscores that table localization remains a key bottleneck. \data{} enables inclusive, structure-aware document AI and supports scalable research on document intelligence.

\section{Limitation}
While \data{} addresses an important gap in multilingual and cross-lingual table-based VQA, our work has several limitations that point to directions for future research.

Our Benchmark is table-centric; it can be further expanded to other layouts, such as charts and histograms, which have not yet been explored. Furthermore, our spatial priors rely on table-level bounding boxes, which improve performance. However, we believe that more fine-grained supervision, such as row, column, or cell-level structure, can further enhance performance, which remains to be explored. Incorporating richer structural annotations could further disentangle visual perception errors from reasoning errors.

\bibliography{custom}

\appendix

\section{Appendix} \label{sec:appendix}
Our dataset contains:
\begin{table}[h]
\centering
\begin{tabular}{|l|c|}
\hline
\textbf{Question Type} & \textbf{Percentage (\%)} \\
\hline
Aggregation      & 58.1 \\
Lookup           & 33.2 \\
Comparison       & 1.6  \\
Table Structure  & 6.9  \\
\hline
\end{tabular}
\caption{Distribution of Question Types}
\end{table}

\subsection{Error Analysis}\label{error}

\begin{figure*}[h]
\centering
\includegraphics[width=0.7\textwidth]{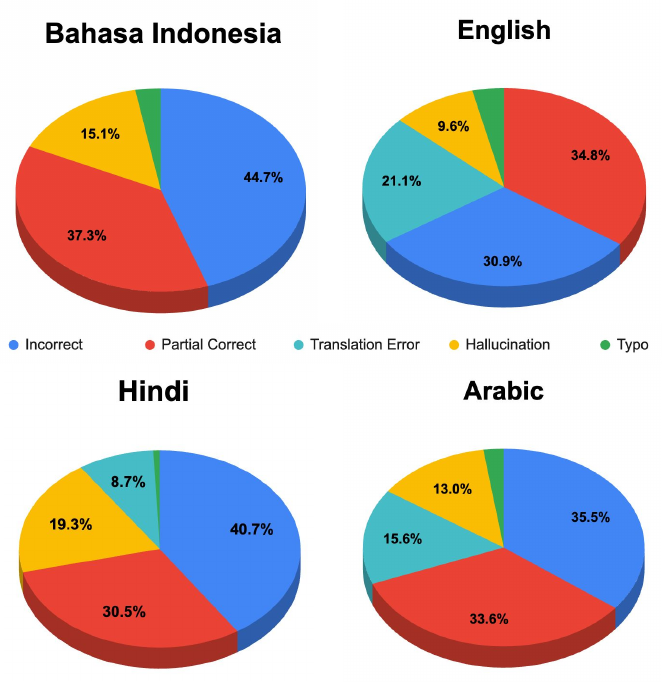}
\caption{Comparative distribution of prediction error types across the four languages in the \data{} test set. The analysis categorizes failures into five types, revealing that `Incorrect' and `Partial Correct' are the dominant error modes. `Translation Error' is a significant factor unique to the cross-lingual settings (English, Hindi, and Arabic), while 'Hallucination' and 'Typo' represent smaller but consistent sources of failure.}
\label{pie_chart}
\end{figure*}

\textbf{Error Taxonomy:} Incorrect (36-45\%): Answer is completely wrong, no semantic relation to ground truth. Hallucination (10-20\%): Model generates plausible but unsupported information. Partial Correct (31-38\%): Answer includes correct information but adds or misses components. Typo (0.7-7\%): Minor lexical variation or spelling error. Translation Error (9-22\%, cross-lingual only): Misunderstanding due to language-specific phrasing.

\textbf{Key Insights from Error Analysis:} Borderless tables produce more hallucinations: Without a clear structure, models are more likely to invent relationships between cells. Cross-lingual errors are often translation-related: Models sometimes respond in the wrong language or misinterpret culture-specific terms. Numerical errors are rare but catastrophic: When models misread numbers, errors are factually wrong (not just semantic variations). Complex tables increase all error types: Tables with merged cells, nested headers, or irregular layouts have 2× higher error rates.

\textbf{Answer Format:} All answers are kept concise (1-5 words typically) and consistent across languages. Numerical answers use standard formatting (e.g., "1,380,201" or "1.380.201" depending on locale conventions).
\subsection{Training Setup}\label{Train_setup}
All experiments were conducted on a single NVIDIA RTX A6000 GPU (48GB), where we fine-tuned Qwen2.5VL-3B using mixed-precision (bfloat16) with gradient checkpointing. Training used the Hugging Face \texttt{Trainer} API with an effective batch size of 4~(per-device batch size of 1 with 4-step gradient accumulation), learning rate of 2e-5, AdamW optimizer, and a linear schedule over 3 epochs, with separate models trained for each of the 4 languages and we used the LoRA method to finetune the Qwen2.5-VL-7B model.

\begin{table}
\scriptsize
\centering
\begin{tabular}{ccccc}
\toprule
\textbf{Benchmark} & \makecell{\textbf{Cross} \\ \textbf{lingual}} & \makecell{\textbf{QA} \\ \textbf{Language}} & \makecell{\textbf{Visual} \\ \textbf{Language}} & \makecell{\textbf{Table}\\\textbf{Focus}}\\
\midrule
\textbf{Tabular VQA:} &  &  &  \\
DocVQA & \xmark  & English & English & Partial\\
TableVQA-B  & \xmark & English & English & \cmark\\
TabComp & \xmark & English & English & \cmark\\
ComTQA & \xmark & English & English & \cmark\\
XT-VQA & \cmark & (EN/FR/CH) & English & Partial\\
MMCricBench & \cmark & English & EN/HI & \cmark\\
\textbf{Ours:} &  &  &  \\
\makecell{\textbf{\data{}} \\ (3B + 7B)} & \cmark & \makecell{4 language \\ (ID/EN/HI/AR)} & \makecell{Bahasa \\ Indonesia} & \cmark\\
\midrule
\end{tabular}
\caption{\textbf{Comparison of VQA Benchmarks:} Most existing benchmarks support the English language for visual content and QA. In contrast, our benchmark focuses on underrepresented low-resource languages, such as Bahasa Indonesia in vision, and includes a variety of languages in QA (including the reading order in case of Arabic). TableVQA-B is TableVQA-Bench~\cite{kim2024tablevqa}}
\label{statistics2}
\end{table}
\subsection{How \data{} is different:}
\begin{itemize}
    \item Geographic and linguistic diversity: We cover Southeast Asia (Bahasa Indonesia), South Asia (Hindi), and the Middle East (Arabic), alongside English, to ensure both regional representation and global accessibility.
    \item Script diversity: We include Devanagari (Hindi) and right-to-left script (Arabic), which pose different challenges than Latin/CJK scripts.
    \item Diverse and Real-world images: Our benchmarks consist of real-world images featuring various styles of table images.
\end{itemize}
\begin{table}[h]
\centering
\scriptsize
\begin{tabular}{l p{4.2cm}}
\toprule
\textbf{Aspect} & \textbf{Protocol Summary} \\
\midrule
Question types & Lookup, aggregation, comparison, structure \\
Avg. tokens per question & 7–10 (across languages) \\
Translation workflow & MT → human validation → consistency check \\
Discarded items & 3.1\% (ambiguity, mistranslation) \\
Annotators & 3 (Bahasa), 4 validators (EN/HI/AR) \\
\bottomrule
\end{tabular}
\caption{Annotation protocol summary}
\label{annotation}
\end{table}
\begin{table*}
\centering
\scriptsize
\begin{tabular}{p{6.6cm}!{\vrule width 1pt}p{2.4cm}!{\vrule width 1pt}p{3.0cm}!{\vrule width 1pt}p{0.4cm}}
\toprule
\textbf{Question} & \textbf{Answer} & \textbf{Predicted answer} & \textbf{E} \\ \midrule
Which column has 4 points as its contents? & form of learning & \textcolor{chartyellow}{day (hour) (j), weight of value} & H \\ \midrule
What is the most common type of disability in Indonesia? & mentally disabled & \textcolor{chartblue}{physical handicap, multiple..} & I \\ \midrule
What is the unit of school that receives the l... & school & \textcolor{chartcyan}{sekolah} & T \\ \midrule
What description is given from the falcon and eagle? & (all types of the family) & \textcolor{chartgreen}{all species from the family} & Ty \\ \midrule
Which province has the highest number of persons with disabilities? & yogyakarta & \textcolor{chartred}{west java, yogyakarta} & P \\
\bottomrule
\end{tabular}
\caption{Examples of errors, H=Hallucination, I=Incorrect, T=Translation, Ty=Typo, P=Partial Correct}
\label{tab:error_table}
\end{table*}
\begin{figure*}
\centering
\scriptsize
\includegraphics[width=\textwidth]{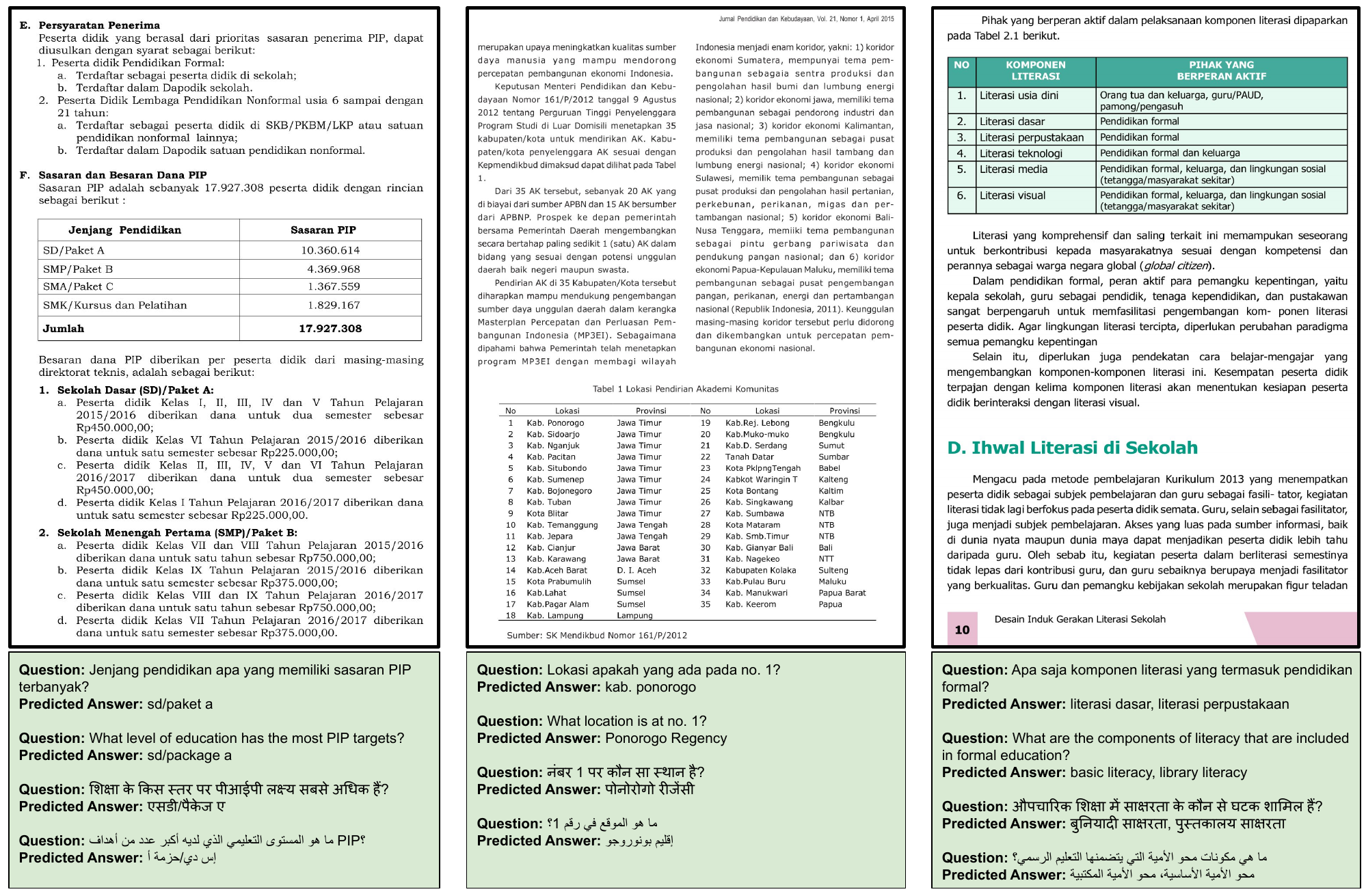}
\caption{Example of the \data{} correct predictions on mono-lingual and cross-lingual question answering across three table formats. Bordered (left), Borderless (middle), and Colorful (right). The examples include questions in Bahasa Indonesia, English, Hindi, and Arabic.}
\label{last_diag}
\end{figure*}

\end{document}